\documentclass{article}

\usepackage[preprint]{neurips_2025}

\usepackage[utf8]{inputenc} 
\usepackage[T1]{fontenc}    
\usepackage{hyperref}       
\usepackage{url}            
\usepackage{booktabs}       
\usepackage{amsfonts}       
\usepackage{nicefrac}       
\usepackage{microtype}      
\usepackage[dvipsnames]{xcolor}
\usepackage{amsmath} 
\usepackage{bm}

\newcommand{\EDGES}{\mathcal{E}} 

\newcommand{\NODES}{\mathcal{N}}
\newcommand{\SG}{S^{\text{g}}} 
\newcommand{\SD}{S^{\text{d}}} 
\newcommand{\V}{\mathbf{V}}  
\newcommand{\VM}{\mathbf{v}} 
\newcommand{\SF}{S^{\text{f}}} 

\usepackage{xcolor,stackengine}
\newcommand\radbox[1]{%
  \fboxsep=2pt
  \fboxrule=.75pt
  \def\tmp{\displaystyle\strut #1}
  \def\shadow{\makebox[0pt]{$\tmp$}}
  \stackengine{0pt}{%
    \stackengine{0pt}{%
      \textcolor{red}{\fbox{$\phantom{\tmp}$}}%
    }{\color{white}}{O}{c}{F}{F}{L}%
  }{$\tmp$}{O}{c}{F}{F}{L}
}

\usepackage{float}
\newfloat{model}{thp}{lop}\floatname{model}{Model}
\usepackage{graphicx}
\usepackage{subcaption}

\usepackage[skins,theorems]{tcolorbox}
\tcbset{highlight math style={enhanced,
  colframe=red,colback=white,arc=0pt,boxrule=1pt}}

\usepackage{mathtools}

\newcommand{\vectorproj}[2][]{\textit{proj}_{\vect{#1}}\vect{#2}}
\newcommand{\vect}{\mathbf}

\newcommand{\R}{\ensuremath{\mathbb{R}}}

\newcommand{\ieee}{\texttt{ieee300}}
\newcommand{\pegaseS}{\texttt{pegase1k}}

\newcommand{\rte}{\texttt{rte6k}}

\usepackage{enumitem}
\setlist{nosep}

\title{Sobolev Training of End-to-End Optimization proxies}

\author{%
 Andrew~W.~Rosemberg \\
 Georgia Institute of Technology\\
 Atlanta, GA, United States \\
 \texttt{arosemberg3@gatech.edu} \\
 \And
 Joaquim Dias Garcia \\
 PSR \\
 Rio de Janeiro, RJ, Brazil \\
 \texttt{joaquim@psr-inc.com} \\
    \And
  Russell Bent\\
 Los Alamos National Laboratory \\
 Los Alamos, NM, US \\
 \texttt{rbent@lanl.gov} \\
  \And
 Pascal Van Hentenryck \\
 Georgia Institute of Technology\\
 Atlanta, GA, United States \\
 \texttt{pvh@gatech.edu} \\
}

\newtheorem{theorem}{Theorem}

\newtheorem{remark}{Remark}
\newtheorem{assumption}{Assumption}
\newtheorem{proof}{Proof}

\begin{document}

\maketitle

\begin{abstract}
Optimization proxies—machine-learning models trained to approximate
the solution mapping of parametric optimization problems in a single
forward pass—offer dramatic reductions in inference time compared to
traditional iterative solvers.  This work investigates the integration
of solver sensitivities into such end-to-end proxies via a
Sobolev–training paradigm and does so in \emph{two distinct settings}:
(i) \emph{fully supervised} proxies, where exact solver outputs and
sensitivities are available, and (ii) \emph{self-supervised} proxies
that rely only on the objective and constraint structure of the
underlying optimization problem.  
By augmenting the standard training loss
with directional-derivative information extracted from the solver, the
proxy aligns both its predicted solutions \emph{and} local derivatives
with those of the optimizer.  Under Lipschitz-continuity
assumptions on the true solution mapping, matching
first-order sensitivities is shown to yield uniform approximation error
proportional to the training-set covering radius.  
Empirically, different impacts are observed in each studied setting.  
On three large Alternating Current
Optimal Power Flow benchmarks, supervised Sobolev training cuts mean-squared error
by up to 56 \% and the median worst-case constraint violation by up to
400 \% while keeping the optimality gap below 0.22 \%.  
For a mean–variance
portfolio task trained without labeled solutions, self-supervised Sobolev training
halves the average optimality gap in the medium-risk region
(i.e. standard deviation above $10\%$ of budget) and matches the baseline elsewhere.
Together these results highlight Sobolev training—whether supervised or
self-supervised—as a path to fast, reliable surrogates for
safety-critical, large-scale optimization workloads.
\end{abstract}

\section{Introduction}\label{sec:intro}

\emph{Optimization proxies} \cite{van2025optimization,
  amos2023tutorial, donti2021dc3} are learned functions that emulate
the solution operator of a parameterized optimization problem,
\[
\begin{aligned}
g(p) 
&= \arg\min_{x}\;\bigl\{\,f(x; p)\;\big|\;c(x; p) = 0,\; x \ge 0\,\bigr\} = x^*, \\[6pt]
\hat g_\theta(p)
&\approx
g(p),
\quad
\hat g_\theta: p \;\mapsto\; \tilde{x} \;\approx\; x^*.
\end{aligned}
\]
thereby replacing an iterative solver with a single forward pass,
where $p\in\mathbb{R}^d$ denotes the problem parameters,
$x\in\mathbb{R}^n$ the decision variables, and $f,c$ are
differentiable functions.  Such surrogates enable millisecond-scale
inference in time-critical domains—real-time grid dispatch, on-device
resource allocation, quantitative finance—and, when embedded inside
larger decision pipelines, they provide differentiable “inner loops’’
that allow an outer optimization to steer itself with respect to
problem parameters via the sensitivities $\nabla_p\hat g_\theta$.
Here, inaccurate sensitivities can lead to slow convergence and poor
performing solutions in such nested workflows.

This study explores the idea of jointly learning solution values with a model architecture that approximates the geometric sensitivities of the problem
 in optimization proxies. It adopts a Sobolev-style loss
that supplements value regression with first-order information, an
idea first explored for generic function approximation in
\cite{czarnecki2017sobolev} and later for approximating functions
embedded inside optimization problems \cite{tsay2021sobolev}.  For
optimization problems, the sensitivity analysis of the solutions can
be automatically extracted based on methods appropriate to the
underlying problem class \cite{pacaud2025sensitivity}. Expanding on
prior literature, the present work makes four main contributions:

\begin{enumerate}[leftmargin=*]
\item \textbf{Scalable, Sparse-Masked Sobolev Training.}
      A sparsity-masked, first-order Sobolev loss enforces agreement between a
      proxy’s Jacobian and solver sensitivities on only a carefully chosen
      subset of partial derivatives.  
      The mask (i) cuts Jacobian memory requirements even further than the
      vector–Jacobian contraction of \cite{czarnecki2017sobolev},
      (ii) eliminates the gradient interference that appears under dense
      supervision, and (iii) stabilizes GPU training on general
      large-scale optimization problems.

\item \textbf{Uniform value–gradient error bounds.}  The paper derives
  multiple approximation bounds, including a bound that is
  proportional to the \emph{square} of the training-set covering
  radius. The bounds extend classical results on value-only matching
  to joint value-and-gradient consistency, under mild Lipschitz
  continuity of the exact solution map and \(C^{2}\) neural
  activations of bounded curvature. This guarantee applies to convex
  problems and non-convex programs satisfying linear independence constraint qualification (LICQ),  second order sufficient conditions (SOSC), and related
  regularity conditions.

\item \textbf{Practical data-generation and implementation.}  The
  paper proposes an end-to-end pipeline that automatically (i)
  produces a broad spectrum of problem instances that cover both
  nominal and edge-case operating conditions, (ii) retrieves exact
  parameter sensitivities by differentiating the solver’s Karush-Kuhn-Tucker (KKT) system,
  (iii) sparsifies the resulting Jacobians with a task-aware mask to
  curb memory and compute demands without sacrificing fidelity, and
  (iv) relies on modern differentiable-programming tooling—implemented
  in Julia but agnostic to language—to train multi-thousand-variable
  proxies comfortably on a single GPU.

\item \textbf{Comprehensive empirical validation.}  The paper discusses the applicability of the method for training optimization surrogates in different contexts. First, 
  on three industry AC-OPF benchmarks -- IEEE-300, PEGASE-1k, and
  RTE-6k \cite{pglib} --, it shows that the Sobolev-trained proxy lowers
  mean-squared error by as much as 56 \% and reduces median worst-case
  constraint violation by up to a factor of four, while keeping the
  relative optimality gap below 0.22 \%.  Second, on a self-supervised study on
  mean–variance portfolio selection, it shows a different pattern: in the
  tight-risk regime \(\sigma_{\max}\le0.10\,\mathcal B\), the benchmark
  that ignores derivatives performs \emph{slightly} better, yet, for looser risk
  budgets, the Sobolev variant cuts the average optimality gap almost
  in half (from \(18.9\pm20.2\%\) to \(8.7\pm9.5\%\).  The pronounced and consistent
  split in both training and test suggests a mixture-of-experts strategy that calls the
  benchmark inside the high-constraint region and the Sobolev proxy
  elsewhere.
\end{enumerate}



Together, the theoretical guarantee and the two contrasting
application domains indicate that Sobolev-based training can endow
optimization proxies with dependable accuracy \emph{and} high-fidelity
gradients, advancing their suitability for safety-critical,
large-scale decision systems.

The remainder of the paper is organized as follows. Section
\ref{sec:background} reviews the sensitivity-analysis literature that
underpins the proposed approach and situates Sobolev training within related
work on differentiable optimization layers.  Section \ref{sec:sobolev}
formalizes the masked Sobolev loss, details the data-generation
pipeline for solver sensitivities, and discusses practical
implementation choices.  Section \ref{sec:theory} establishes uniform
value-and-gradient error bounds and specifies the conditions under
which they hold.  Section \ref{sec:opf} reports supervised results on
three large AC-OPF benchmarks, while Section \ref{sec:case-portfolio}
presents a self-supervised study on mean–variance portfolio selection
and motivates a mixture-of-experts extension.  Section
\ref{sec:limitations} summarizes observed limitations and open
challenges, and Section \ref{sec:conclusion} concludes with avenues
for future research.

\section{Background and Literature Review}
\label{sec:background}

Early investigations into the differentiability of solutions to
constrained optimization problems can be traced back to
\cite{fiacco1976sensitivity}, which established conditions under which
a KKT system exhibits stable, differentiable
solutions. In particular, that work emphasized the importance of
having a unique primal solution mapping, a unique dual solution, and
local stability of active constraints. Under assumptions such as
SOSC, LICQ, and strict complementarity slackness (SCS),
these KKT conditions form a set of smooth equations involving both
decision variables and parameters, allowing the application of the
implicit function theorem to determine how optimal solutions vary with
respect to parameter changes.

Subsequent efforts relaxed some of these strict requirements to
address degeneracies. For instance, \cite{kojima1980} developed
concepts like Mangasarian-Fromovitz Constraint Qualification (MFCQ)
 and Generalized Strong Second-Order Sufficient Condition (GSSOSC), which help preserve stability without
imposing strict complementarity. This line of research also includes
the works of \cite{jittorntrum1984} and \cite{shapiro1985}, who
explored the notion of directional differentiability in the absence of
LICQ, and \cite{ralph1995}, which provided practical methods for
evaluating directional derivatives. Over time, sensitivity analysis
expanded to more complex frameworks such as variational inequalities,
bi-level formulations \cite{dempe2002}, stochastic programs
\cite{shapiro1990,shapiro1991}, and model predictive control
\cite{zavala2009,jin2014}, culminating in the creation of robust
software implementations \cite{pirnay2012optimal,andersson2019casadi}.

Meanwhile, the domain of gradient-based optimization and
differentiable programming gained substantial momentum in machine
learning
\cite{huangfu2018parallelizing,shin2023accelerating,lubin2023jump,innes2019differentiable}. A
significant outcome of this trend is the integration of constrained
optimization methods directly into neural networks, which enables
end-to-end training pipelines that merge data-driven layers with
principled decision models \cite{amos2017optnet,
  DeepDeclarativeNetworks}. Several toolkits emerged to
streamline these capabilities, including CVXPY-Layers
\cite{agrawal2019differentiable}, sIPOPT \cite{pirnay2012optimal},
CasADi \cite{andersson2019casadi}, and Theseus
\cite{pineda2022theseus}, all of which provide interfaces for treating
solvers as differentiable modules. In the Julia ecosystem,
\texttt{DiffOpt.jl} \cite{besanccon2024flexible} has extended
\texttt{JuMP.jl} to offer solution sensitivities for convex models,
with ongoing development targeted at broader non-convex settings.

\vspace{-0.3cm}
\subsection{Sensitivity Calculation for Parametric Problems}

Let $\mathbf{s}^* = \bigl(x(p),\lambda(p)\bigr)$ be a local
solution consisting of primal and dual (Lagrange multiplier)
variables that satisfy the KKT conditions: $F\bigl(\mathbf{s}^*,
p\bigr) \;=\; 0$, where $F$ encapsulates feasibility, stationarity, and
complementary slackness in a single system of equations. Under
appropriate regularity assumptions (e.g., LICQ, SOSC, strict
complementarity) ensuring a unique local optimum and non-singular
derivative $\nabla_{\mathbf{s}}F(\mathbf{s}^*,p)$, one may apply the
implicit function theorem to obtain:
\begin{equation}\label{eq:kkt_sensitivity}
\nabla_{p} \, \mathbf{s}^*
=
-\,\Bigl(\nabla_{\mathbf{s}}F(\mathbf{s}^*,p)\Bigr)^{-1}
\;\nabla_{p}F(\mathbf{s}^*,p).
\end{equation}
This expression reveals how small changes in the parameter vector $p$
propagate through the KKT system to induce changes in both $x(p)$ and
its dual variables $\lambda(p)$. In practice, computing these
derivatives typically involves linear algebra on the Jacobian of the
KKT system. The sensitivity information is valuable for a variety of
tasks, including stability analysis, local robustness assessments, and
(as studied in this work) using partial derivatives to guide the
training of models that approximate the solution mapping $p \mapsto
x^*(p)$.

\vspace{-0.3cm}
\section{Methodology}\label{sec:sobolev}

\begin{figure}[t]
  \centering
  \includegraphics[width=0.8\textwidth]{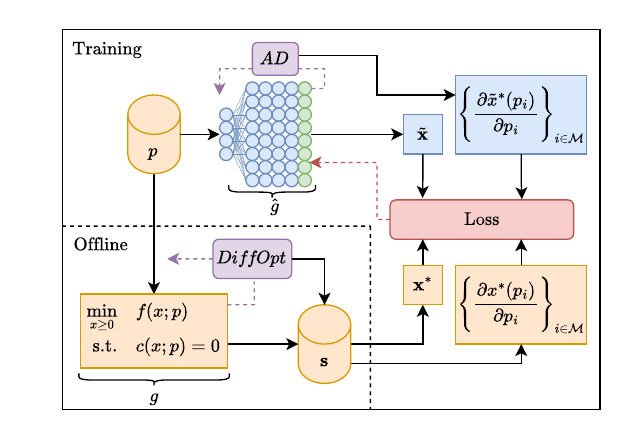}
  \vspace{-0.2cm}
  \caption{End-to-end pipeline for Sobolev training of optimization
           proxies.  The {\color{MidnightBlue}{upper}} half shows the
           online training loop driven by automatic differentiation
           (AD); the {\color{BurntOrange}{lower}} half is the offline
           oracle that supplies ground-truth solutions~$x^\star$ and
           (masked) sensitivities via a Differentiable Optimization (\textsc{DiffOpt}) layer.  Solid arrows
           are forward evaluations, dashed arrows are
           Jacobian/Gradient calculations.}
  \label{fig:diagram_sobolev}
\end{figure}

\paragraph{Sobolev Training.}
Sobolev Training augments conventional regression by additionally
matching prescribed sensitivities. As sketched in the \emph{{\color{MidnightBlue}{upper}} half of Fig.~\ref{fig:diagram_sobolev}},
given tuples \( (p_i,\, x^* = g(p_i),\,
Dg(p_i)=\partial x^\star/\partial p_i)\), a first-order (masked) Sobolev {\color{Red}{Loss}} is defined as
\[
\mathcal{L}(\theta)=
\frac{1}{N}\sum_{i=1}^{N}
      \ell\bigl(\hat{g}_\theta(p_i),\,g(p_i)\bigr)
\;+\;
\frac{\lambda}{N}\sum_{i=1}^{N}
        \ell_d\bigl(\mathcal{M}\odot D\hat{g}_\theta(p_i),\;
                       \mathcal{M}\odot Dg(p_i)\bigr),
\]
where \(D\hat{g}_\theta(p_i)\) denotes the Jacobian of the network at
\(p_i\) - provided by the same Automatic Differentiation framework ({\color{Purple}{AD}}) already used for getting gradients of the loss w.r.t. network inputs; \(\ell\) and \(\ell_d\) measure discrepancies on outputs and
directional derivatives respectively, \(\lambda\) balances the two
terms, and \(\mathcal{M}\in\{0,1\}^{d\times d}\) is a binary mask that through the element wise operation, $\odot$, selects the subset of sensitivities under consideration. 

\vspace{-0.2cm}
\paragraph{Sparse Masking.} 

The masking proposed in this paper
helps mitigate the memory cost of full-Jacobian matching and
alleviates gradient conflicts reported in large-scale constrained
learning \cite{liu2024config}. Empirical results show that extreme sparsity—masking between 75 \% and 95 \% of Jacobian entries (i.e.\ keeping only 5–25 \%)—yields the most stable training and lowest test‐set MSE on large surrogate models.  An ablative study on the IEEE-300 case, detailing how performance varies with mask density, is provided in Appendix \ref{app:mask}.

\vspace{-0.3cm}
\paragraph{Supervised vs Self-Supervised}

The expression \(\ell\) can act as a purely supervised regression loss
when ground-truth solutions and their derivatives are available, exploiting the
\emph{optimization model itself} as a teacher. In
situations where exact solutions are unknown, expensive or may
represent a prohibitive hard mapping, the optimization
objective regularized by penalties for constraint
violation (also know as the Lagrangian function) provides an unsupervised signal:
\(
    \tilde f(x;p)
      := f(x;p)\;+\;\beta\,\|c(x;p)\|_2^2
      +\;\gamma\,\bigl\langle x\bigr\rangle_{-},
\)
with $\beta,\gamma\!>\!0$ and $\langle \cdot \rangle_{-}$ denoting the
parts of the $x$ not respecting the non-negativity constraint.

\vspace{-0.2cm}
\paragraph{Feasibility-Preserving Proxies}

While the training of optimization proxies also aims at reducing infeasibility, the raw output predicted by the neural network, $\hat g_\theta : p \rightarrow \tilde{x}$, is not required to satisfy the problem constraints:
\(
    \mathcal{C}(p)\;=\;\{\,x\in\mathbb{R}^n \mid c(x;p)=0,\; x \geq 0\,\}.
\)
In situations where a projection, $\vectorproj[\mathcal{C}(p)]{\tilde{x}}$, is easily computed and has (at least) directional derivatives,  \emph{feasibility-enforcing layers} (e.g. {\color{Green}{last}} network layer in  Fig.~\ref{fig:diagram_sobolev}) have been shown to improve performance. This is the case for one of the applications in the present study presented in Section \ref{sec:case-portfolio}.

\vspace{-0.2cm}
\paragraph{Dataset Creation.}

Constructing a dataset that faithfully represents the solver’s
behavior across both feasible and infeasible regions requires a blend
of sampling strategies.  Uniform ``box’’ sampling draws within
believed parameter feasible bounds, while incremental ``line’’
excursions along single coordinates expose the boundary limits of
feasibility. Additionally, realistic distributions over parameters,
can be used when they are known (as in the case of Power
Systems). Combining these three procedures in fixed proportions yields
a dataset that spans the operating domain without excessive
redundancy.

In Sobolev training, where a supervised regression loss is needed for
at least matching sensitivities, the workflow proceeds as follows.  An
optimization instance is solved for every parameter realization
\(p_i\) in the training set with an appropriate solver such as \textsc{Ipopt} \cite{ipopt} ({\color{BurntOrange}{Optimization}} block in Fig.~\ref{fig:diagram_sobolev})
to obtain the primal (sometimes local) optimum \(x^\star(p_i)\).  Solver
sensitivities \(Dg(p_i)\) are extracted
through sensitivity analysis using a software such as
{\color{Purple}{\texttt{DiffOpt.jl}}}, and finally a sparse mask \(\mathcal{M}\) is
randomly sampled for each instance. Once the dataset is created \footnote{If one has already stored the primal and dual solutions and the problem structure is available, there is no need to re-solve to compute derivatives.}, the Sobolev training
loop proceeds much like standard training. The \emph{{\color{BurntOrange}{lower}} half of Fig.~\ref{fig:diagram_sobolev}} sketches the order of the different procedures in the offline dataset creation and how it feeds into training.


\vspace{-0.3cm}
\section{Theoretical Perspective}
\label{sec:theory}

Many classical universal approximation theorems show that neural
networks can approximate continuous functions arbitrarily well in
$L^2-$ or uniform norms. However, matching derivatives requires
approximation in a Sobolev norm. Earlier results
\cite{czarnecki2017sobolev} show that standard networks with
sufficiently rich activation functions can also approximate
derivatives.

One key advantage of incorporating derivatives is the reduction in
sample complexity. In certain families of functions, it can take fewer
input-output-derivative samples to identify a target function than
input-output pairs alone. Indeed, the training set conveys, not only
the function value at a point, but also how it changes locally,
reducing the degrees of freedom in the hypothesis space. As in
classical polynomial-fitting arguments, knowledge of derivatives at a
point can disambiguate many potential fits. Recent results have
demonstrated how approximating derivatives (either explicitly or
implicitly) \cite{Zhang2022_ConvexNNSolverDCOPF,
  rosemberg2024learning} can help achieve small generalization errors
in the test set.  This effect becomes especially significant in
high-dimensional problems or in the presence of strong
non-linearities, where matching values alone might fail to capture the
local geometry of \( f \). Moreover, classical results from
sensitivity analysis (e.g., \cite{liu1995sensitivity}) guarantee that
for a large class of convex and well-behaved nonconvex optimization
problems, the optimal solution mapping is continuous or even
Lipschitz-continuous. In particular, if the optimization problem is
convex and satisfies regularity conditions such as Slater’s condition,
then small perturbations in the problem parameters lead only to
bounded—and typically linear—changes in the optimal
solution. Consequently, the derivatives (when they exist) are finite
and well-behaved, ensuring that a Sobolev-trained model which matches
both the values and sensitivities at training points can interpolate
accurately between them. This bounded, regular behavior of the
solution mapping is precisely the scenario where Sobolev Training
excels, as it can exploit these local regularity properties to greatly
reduce sample complexity and improve generalization.

\noindent  The ensuing three approximation theorems formalize this intuition, bounding the proxy’s uniform error under value-only, Jacobian-only, and joint Sobolev matching.

Let \(\mathcal P\subset\R^{d}\) be a compact parameter space and
\(\mathcal T=\{p_i\}_{i=1}^{N}\subset\mathcal P\) a finite training set whose
 \emph{covering radius} is
\[
\delta\;:=\;\sup_{p\in\mathcal P}\min_{1\le i\le N}\|p-p_i\|.
\]

Denote the exact solution operator by
\(g:\mathcal P\to\R^{n}\) and the learned proxy by
\(\hat g_\theta:\mathcal P\to\R^{n}\).  Sobolev training enforces
\[
\hat g_\theta(p_i)=g(p_i),\qquad
D\hat g_\theta(p_i)=Dg(p_i),\qquad\forall\,p_i\in\mathcal T.
\]

\begin{assumption}[Value-Lipschitz]\label{as:value_lip}
There exist inherent values \(L_g,L_{\hat g}>0\) such that
\(\|g(p)-g(q)\|\le L_g\|p-q\|\) and
\(\|\hat g_\theta(p)-\hat g_\theta(q)\|\le L_{\hat g}\|p-q\|\)
for all \(p,q\in\mathcal P\).
\end{assumption}

\begin{assumption}[Derivative-Lipschitz]\label{as:deriv_lip}
There exist \(M_g,M_{\hat g}>0\) with \(g,\hat g_\theta\in C^{1}(\mathcal P)\) and
\(\|Dg(p)-Dg(q)\|\le M_g\|p-q\|\),
\(\|D\hat g_\theta(p)-D\hat g_\theta(q)\|\le M_{\hat g}\|p-q\|\)
for all \(p,q\in\mathcal P\).
\end{assumption}

\begin{theorem}[Value Matching Only]\label{thm:value_only}
Let $\delta$ be the \emph{covering radius} of $\mathcal{P}$. Under Assumption \ref{as:value_lip},
\[
\sup_{p\in\mathcal P}\bigl\|\hat g_\theta(p)-g(p)\bigr\|
\;\le\;(L_g+L_{\hat g})\,\delta.
\]
\end{theorem}

\begin{proof}
Fix any \(p\in\mathcal P\) and choose \(p_j\in\mathcal T\) with
\(\|p-p_j\|\le\delta\).  By the triangle inequality and Lipschitz bounds,
\[
\begin{aligned}
\|\hat g_\theta(p)-g(p)\|
&\le \|\hat g_\theta(p)-\hat g_\theta(p_j)\|
     +\|g(p_j)-g(p)\|   \\[2pt]
&\le L_{\hat g}\|p-p_j\|+L_g\|p-p_j\|
\;\le\;(L_g+L_{\hat g})\,\delta.
\end{aligned}
\]
Taking the supremum over \(p\) completes the proof.
\end{proof}

\begin{theorem}[Jacobian Matching Only]\label{thm:deriv_only}
Let $\delta$ be the \emph{covering radius} of $\mathcal{P}$. Under Assumption \ref{as:deriv_lip},
\[
\sup_{p\in\mathcal P}\bigl\|D\hat g_\theta(p)-Dg(p)\bigr\|
\;\le\;(M_g+M_{\hat g})\,\delta.
\]
\end{theorem}

\begin{proof}
Repeat the argument above with \(Dg\) and \(D\hat g_\theta\) in place of
\(g\) and \(\hat g_\theta\) and with \(M\)-constants instead of \(L\)-constants.
\end{proof}

\begin{theorem}[Sobolev Guarantee]\label{thm:sobolev}
Let $\delta$ be the \emph{covering radius} of $\mathcal{P}$. Under Assumptions \ref{as:deriv_lip} and with both value
and Jacobian interpolation on \(\mathcal T\),
\[
\sup_{p\in\mathcal P}\bigl\|\hat g_\theta(p)-g(p)\bigr\|
\;\le\;\tfrac12\,(M_g+M_{\hat g})\,\delta^{2}.
\]
\end{theorem}

\begin{proof}
Let \(p\in\mathcal P\) be arbitrary and pick its nearest training point
\(p_j\) so that \(h:=p-p_j\) satisfies \(\|h\|\le\delta\).
For any \(C^{1}\) map \(f\) the fundamental theorem of calculus gives
\[
f(p)=f(p_j)+Df(p_j)h+\int_{0}^{1}\!\bigl(Df(p_j+th)-Df(p_j)\bigr)h\,dt.
\]
Define the remainder \(R_f(h):=\int_{0}^{1}\bigl(Df(p_j+th)-Df(p_j)\bigr)h\,dt\).
Derivative-Lipschitz continuity yields
\(\|R_f(h)\|\le\frac12 M_f\|h\|^{2}\).  
Applying this to \(f=g\) and \(f=\hat g_\theta\) and using
exact interpolation at \(p_j\) cancels the zeroth- and first-order terms,
so
\[
\hat g_\theta(p)-g(p)=R_{\hat g_\theta}(h)-R_g(h)
\quad\Longrightarrow\quad
\|\hat g_\theta(p)-g(p)\|
\le\tfrac12(M_g+M_{\hat g})\,\|h\|^{2}
\le\tfrac12(M_g+M_{\hat g})\,\delta^{2}.
\]
\end{proof}

\begin{remark}
Compactness of \(\mathcal P\) plus standard regularity conditions
(LICQ, SOSC, strict complementarity) guarantee \(g\in C^{1,1}(\mathcal P)\)
with bounded Jacobian, so the assumptions are generically met.
Feed-forward networks whose activations are \(C^{2}\) with bounded curvature
(e.g.\ \(\tanh\), softplus) satisfy the same properties on
compact domains, hence Theorems \ref{thm:value_only}–\ref{thm:sobolev}
apply directly to the proxies considered in this paper.
\end{remark}

\section{Supervised Learning Application: Optimal Power Flow}
\label{sec:opf}

Power-system optimization has surged in importance due to network expansion, deep renewable-energy penetration, and sustainability targets \cite{s2022, pozo2012three, pozo2012chance, barry2022risk}. The central computational task, the AC Optimal Power Flow (AC-OPF) problem, minimizes total generation cost via the objective \eqref{model:acopf:obj} while satisfying Kirchhoff’s current-law balance at every bus \eqref{model:acopf:kirchhoff}, Ohm’s-law branch relations \eqref{model:acopf:ohm:fr}–\eqref{model:acopf:ohm:to}, voltage-magnitude limits \eqref{model:acopf:vmbound}, generator operating bounds \eqref{model:acopf:genbound}, and thermal capacity constraints on apparent power flows \eqref{model:acopf:thrmbound}. High-accuracy AC-OPF solutions enable real-time risk-aware market clearing \cite{tam2011real,e2elr}, day-ahead security-constrained unit commitment \cite{sun2017decomposition}, transmission switching optimization \cite{OTS}, and long-term expansion planning \cite{verma2016transmission}. The concise formulation in Model \ref{model:acopf} employs complex voltages, injections, demands, and branch flows, omitting detailed bus-shunt and transformer models for brevity, and follows the reference implementation in \texttt{PowerModels.jl} \cite{Coffrin2018_PowerModels}. The reference model notation is detailed in Appendix \ref{app:notation}.

\begin{model}[!t]
\vspace{-0.3cm}
  \caption{AC Optimal Power Flow (AC-OPF) - Notations in Appendix \ref{app:notation}. Problem Parameters Emphasized in Red Box.}
  \label{model:acopf}
  \small
  \begin{subequations}
      \begin{align}
          \min_{\SG, \SF, \V} \quad
              & \sum_{i \in \NODES} c_i (\SG_i) \label{model:acopf:obj} \\
          \text{s.t.} \quad
          & \SG_i - \radbox{\SD_i} = \sum_{(i,j) \in \mathcal{E}_{i} \cup \mathcal{E}^{R}_{i}} \SF_{ij}
              & \forall i \in \NODES \label{model:acopf:kirchhoff} \\
          & \SF_{ij} = (Y_{ij} {+} Y^{c}_{ij})^{\star} \V_{i} \V_{i}^{\star} - Y_{ij}^{\star} \V_{i} \V_{j}^{\star}
              & \forall (i,\, j) \in \EDGES \label{model:acopf:ohm:fr} \\
          & \SF_{ji} = (Y_{ij} {+} Y^{c}_{ji})^{\star} \V_{j} \V_{j}^{\star} - Y_{ij}^{\star} \V_{i}^{\star} \V_{j}
              & \forall (i,\, j) \in \EDGES \label{model:acopf:ohm:to} \\
          & \underline{\VM_i} \leq |\V_i| \leq \overline{\VM_i}
              & \forall i \in \NODES
              \label{model:acopf:vmbound} \\
          & \underline{\SG_i} \leq \SG_i \leq \overline{\SG_i}
              & \forall i \in \NODES
              \label{model:acopf:genbound} \\
          & |\SF_{ij} |, |\SF_{ji}| \leq \overline{S_{ij}}^2
              & \forall (i,\, j) \in \EDGES
              \label{model:acopf:thrmbound}
      \end{align}
  \end{subequations}
\end{model}

The nonconvexity and nonlinear physics in AC-OPF lead to poor scaling
of solution time as network size and scenario count grow, often
precluding its direct use in large‐scale or highly uncertain settings
\cite{o2011recent}.  However, the need to solve many similar OPF
instances has spurred research on learning a parametric surrogate
\(p\mapsto x^*(p)\) that delivers near-optimal solutions in a single
forward pass, thereby bypassing expensive iterative algorithms
\cite{park2023self, chatzos2020high} for each parameter instance
$p=\SD$.  One can train a neural network to predict $x^*(p)$ from $p$
using the standard MSE or similar loss, but Sobolev training augments
this with a derivative penalty that aligns $\frac{\partial
  \hat{x}_\theta}{\partial p}$ with $\frac{\partial x^*(p)}{\partial
  p}$.

\subsection{Experimental Setting}
\label{sec:results:setup}

Two neural‐network proxy models were evaluated on the AC‐OPF problem
in Model~\ref{model:acopf}: one trained using a standard MSE loss
(hereafter refereed to as Benchmark/Bench), and another trained with
an additional Sobolev loss term that leverages derivative information
(hereafter refereed to as Sobolev).  Both models employed the same
fully connected architecture with a single hidden layer of width 320,
trained over approximately $10K$ samples for $20K$ epochs. The
validation and test set comprised of around $5K$ problem instances
each. The evaluation uses three test cases from PGLib
\cite{babaeinejadsarookolaee2019power} with up to 6468 buses.  Table
\ref{table: cases} reports, for each system, the number of buses
$|\mathcal{N}|$, branches $|\mathcal{E}|$ and generators
$|\mathcal{G}|$, as well as the nominal total demand
(${\text{P}}^{d}_{\text{ref}}$) and its range across the dataset
($[\underline{\text{P}}^{d}, \bar{\text{P}}^{d}]$). Implementation
details are provided in Appendix \ref{app:imp}.

    \begin{table}[!t]
        \centering
        \caption{Statistics of the PGLib test cases.}
        \label{table: cases}
        \begin{tabular}{lrrrrr}
            \toprule
            \multicolumn{1}{c}{System}
            & \multicolumn{1}{c}{\textbf{$|\mathcal{N}|$}}  
            & \multicolumn{1}{c}{\textbf{$|\mathcal{E}|$}}  
            & \multicolumn{1}{c}{\textbf{$|\mathcal{G}|$}}
            & \multicolumn{1}{c}{${\text{P}}^{d}_{\text{ref}}$}
            & \multicolumn{1}{c}{$[\underline{\text{P}}^{d}, \bar{\text{P}}^{d}]$}
            \\
            \midrule
            \ieee
                & 300  
                & 411  
                & 69   
                & 263
                & [\phantom{0}210, \phantom{0}280]
            \\
            \pegaseS
                & 1354  
                & 1991  
                & 260     
                & 781
                & [\phantom{0}625, \phantom{0}820] 
            \\
            \rte 
                & 6468  
                & 9000  
                & 399      
                & 1109
                & [\phantom{0}887, 1164]
            \\
            \bottomrule
        \end{tabular}
    \end{table}

\subsection{Metrics}
\label{sec:results:metrics}

Three metrics are employed to assess both solution accuracy and
feasibility.
Denote \(\tilde x\) the proxy’s prediction, \(x^*\) the true solution, \(\tilde z=f(\tilde x)\),
and \(z^*=f(x^*)\):
\[
\begin{aligned}
\text{Mean Squared Error: }\mathrm{MSE} &= \frac1n\|\tilde x - x^*\|_2^2,\\
\text{Optimality Gap: }\mathrm{GAP} &= \frac{\lvert \tilde z - z^* \rvert}{\lvert z^*\rvert},\\
\text{Absolute Infeasibility: }\mathrm{INF} &= \frac1{|\mathcal E|+|\mathcal I|}\sum_{\substack{c\in\mathcal E\\g\in\mathcal I}}
  \begin{cases}
    |\,c(\hat x)\,|, & c(x)=0\ \text{(equality)},\\
    \max\{\,g(\hat x),0\}, & g(x)\le0\ \text{(inequality)\,.}
  \end{cases}
\end{aligned}
\]

\subsection{Experimental Results}

\begin{figure}[!t]
  \centering
  \begin{subfigure}[t]{0.4\textwidth}
    \vspace{0pt}    
    \centering
    \begin{tabular}{llcc}
      \toprule
      \textbf{Case}   & \textbf{Metric}        & \textbf{Sobolev}  & \textbf{Bench}   \\
      \midrule
      \ieee           & MSE                    & \textbf{0.0065}   & 0.0070         \\
      \ieee           & GAP                    & 0.22\%            & \textbf{0.02\%}\\
      \ieee           & INF                    & \textbf{0.0099}   & 0.0116         \\
      \midrule
      \pegaseS        & MSE                    & \textbf{0.0072}   & 0.0089         \\
      \pegaseS        & GAP                    & 0.15\%            & \textbf{0.07\%}\\
      \pegaseS        & INF                    & \textbf{0.0066}   & 0.0098         \\
      \midrule
      \rte            & MSE                    & \textbf{0.0004}   & 0.0009         \\
      \rte            & GAP                    & \textbf{0.07\%}   & \textbf{0.07\%}\\
      \rte            & INF                    & \textbf{0.0042}   & 0.0059         \\
      \bottomrule
    \end{tabular}
    \caption{Average performance of OPF Proxy Models (Sobolev vs MSE).}
    \label{tab:opf_1354}
  \end{subfigure}\hfill
  \begin{subfigure}[t]{0.55\textwidth}
    \vspace{0pt}    
    \centering
    \includegraphics[width=\textwidth]{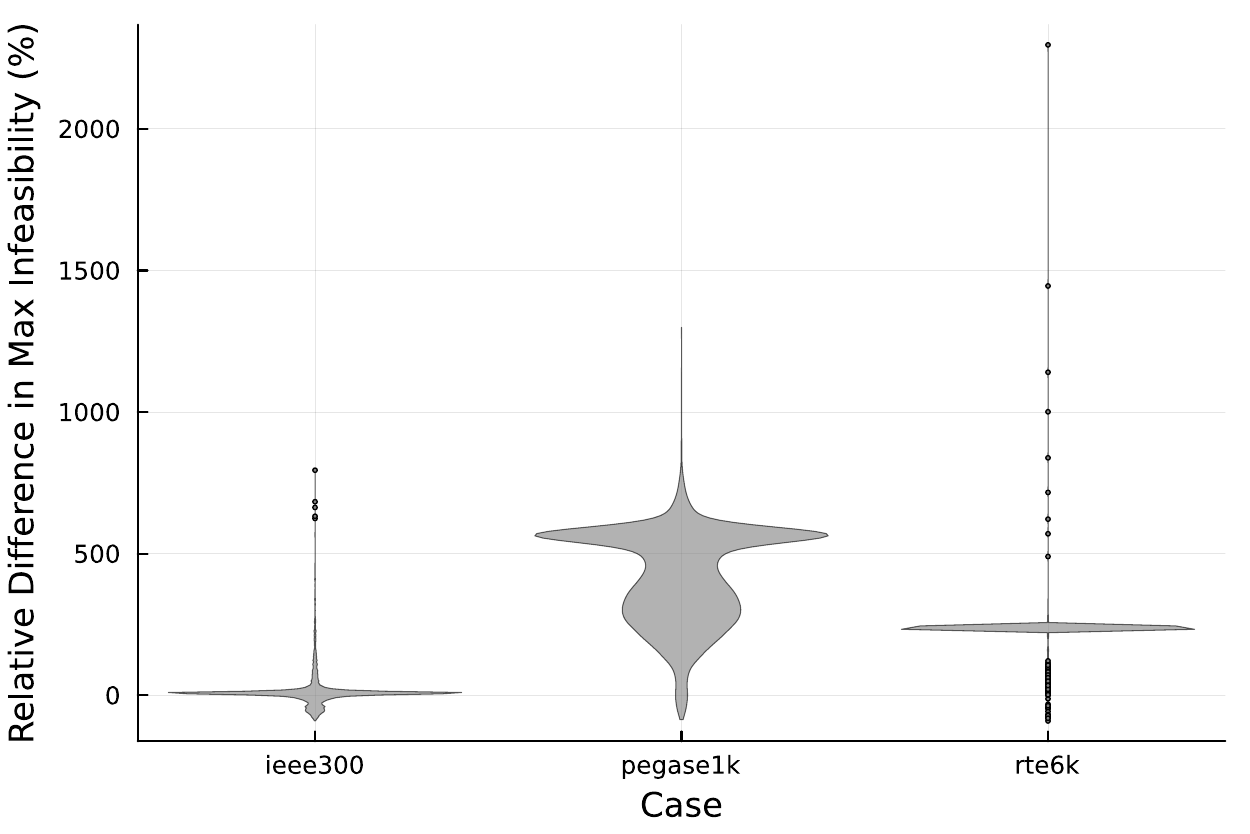}
    \caption{Relative max Infeasibility Difference (\%).}
    \label{fig:all_cases_diff_max_violin}
  \end{subfigure}
  \caption{Comparison of NN Sobolev vs Benchmark across the Three Test Xases.}
  \label{fig:combined}
\end{figure}


Table~\ref{tab:opf_1354} reports the average performance of both
models on the held‐out test sets. The Sobolev‐trained proxy achieves
lower MSE and average absolute infeasibility across all three network
cases. Not surprisingly since it has fewer degrees of freedom, Sobolev
incurs a slightly higher optimality gap on IEEE‐300 and PEGASE‐1k
(0.22\% vs.\ 0.02\%, and 0.15\% vs.\ 0.07\%, respectively); on RTE‐6k
both models match at 0.07\%.


Moreover, the frequency with which higher constraint violations
occurred across test instances was further
examined. Figure~\ref{fig:all_cases_diff_max_violin} presents violin
plots of the relative reduction in maximum infeasibility (RMI) under
Sobolev training, defined for instance \(i\) as
\[
\mathrm{RMI_i} = 100\%\times\frac{\mathrm{max_j}(\mathrm{Infeas}_{\mathrm{MSE_{i,j}}})-\mathrm{max_j}(\mathrm{Infeas}_{\mathrm{Sobolev_{i,j}}})}{\mathrm{max_{i,j}}(\mathrm{Infeas}_{\mathrm{Sobolev_{i,j}}})}.
\]
\noindent
In this definition, $\mathrm{Infeas}_{\mathrm{MSE},i,j}$ and $\mathrm{Infeas}_{\mathrm{Sobolev},i,j}$ are the absolute violations of constraint $j$ on instance $i$ for the MSE‐only and Sobolev‐trained models, respectively. The numerator measures how much the worst‐case violation on instance $i$ is reduced by Sobolev training, and the denominator normalizes by the single largest violation observed across all test instances and constraints (to avoid instances with zero violation). Hence, a positive $\mathrm{RMI}_i$ indicates the percentage by which Sobolev loss decreases the maximum infeasibility relative to the MSE benchmark.

The Sobolev‐based model rarely yields worse violations than the MSE‐only benchmark (fewer than 15\% of cases). Median improvements in maximum infeasibility are approximately 18\%, 400\%, and 180\% for IEEE‐300, PEGASE‐1k, and RTE‐6k, respectively; moreover, Sobolev training substantially suppresses the extreme outliers observed with MSE loss. Around 90\% of the remaining infeasibility originates from minor power‐flow mismatches (\ref{model:acopf:kirchhoff}–\ref{model:acopf:ohm:to}), suggesting that coupling with feasibility‐focused losses or projection techniques could yield further gains.

Overall, while Sobolev loss incurs a small trade‐off in optimality gap (e.g., increasing from 0.02\% to 0.22\% in IEEE‐300 and from 0.07\% to 0.15\% in PEGASE‐1k), the dramatic reduction in average and worst‐case constraint violations makes the Sobolev‐trained proxy significantly more reliable for safety‐critical OPF applications.

\vspace{-0.2cm}
\section{(Semi) Self-Supervised Case Study: Mean–Variance Portfolio Optimization}
\label{sec:case-portfolio}

\vspace{-0.3cm}
Self-Supervised Learning (SSL) of optimization proxies has great benefits since it only requires a distribution of the input data and avoids the need for optimal solutions. Moreover, on some applications,  SSL becomes a necessity as supervised learning may fail to outperform even basic strategies. This happens for the classical Markowitz portfolio-selection (MPS) problem which seeks a weight vector $x\in\mathbb{R}^n$ for $n$ tradable assets that maximizes expected return subject to
a risk budget:
\[
\max_{x \succeq 0} \quad \radbox{\mu}^\top x
\quad\text{s.t.}\quad
x^\top \Sigma x \;\le\; \radbox{\sigma_{\max}^2}, \quad
\mathbf{1}^\top x \leq \mathcal{B},
\]
Here $\mu$ denotes expected returns, $\Sigma$ the covariance matrix, and
$\mathcal B$ the available capital.
The quadratic risk constraint can be rewritten as the second-order cone
\(
\|\,\Sigma^{1/2}x\|_2 \le \sigma_{\max},
\)
where $\Sigma^{1/2}$ is any Cholesky factor and $\sigma_{\max}$ limits portfolio
standard deviation.  The problem is convex, ensuring global optimality, yet solving it to completion may be infeasible in sub-second trading loops when
$n$ is large. 

\vspace{-0.3cm}
\paragraph{Motivation for a proxy.}
Covariances are usually estimated over long horizons and updated
infrequently.  In low-latency settings $\Sigma$ can therefore be considered
fixed, and the optimization treated as parameterized by the rapidly evolving
signals $(\mu,\sigma_{\max})$.  A neural proxy that instantaneously maps these
signals to an approximate solution
$\,x^\star(\mu,\sigma_{\max})$ enables portfolio re-balancing at the pace
required for high-frequency trading.

\vspace{-0.3cm}
\paragraph{Training strategy.}
Supervised learning for MPS resulted in poor proxies with over 95\% optimality gaps, as detailed in Appendix \ref{app:supervised-portfolio}. The results in this section reports on the semi SSL approach. Feasibility can be enforced through inexpensive projection steps
\cite{qiu2024dual}.  Regularity conditions needed for reliable solver
sensitivities hold, making the task suitable for \emph{Sobolev training},
which augments a value-matching loss with a Jacobian term.
A practical complication is sparsity: typically fewer than 15 \% of assets
receive allocations above 0.01 \% of budget, creating a difficult mapping for
purely supervised learners.
Two proxy networks equipped with a feasibility layer are therefore compared: \textbf{Benchmark} which is trained only on objective values; and \textbf{Sobolev} which is trained on values \emph{and} solver sensitivities.

\vspace{-0.3cm}
\paragraph{Results.}
\begin{figure}[!t]
  \centering
  \begin{minipage}[t]{0.4\textwidth}
    \centering
    \vspace{-2cm}
    {\footnotesize%
    \setlength{\tabcolsep}{3.5pt}
    \begin{tabular}{lcc}
      \toprule
      $\sigma_{\max}$ & Sobolev & Bench \\ \midrule
      $\le10\%$ & $29.5\pm26.9$ & $14.7\pm21.2$ \\
      $>10\%$   & $\,8.7\pm\,9.5$ & $18.9\pm20.2$ \\ \bottomrule
    \end{tabular}}
    \vspace{0.5cm}
    \captionof{table}{Test-set gap (mean $\pm$ std)}
    \label{tab:gap-test-stats}
  \end{minipage}\hfill%
  \begin{minipage}[t]{0.6\textwidth}
    \centering
    \includegraphics[width=\linewidth]{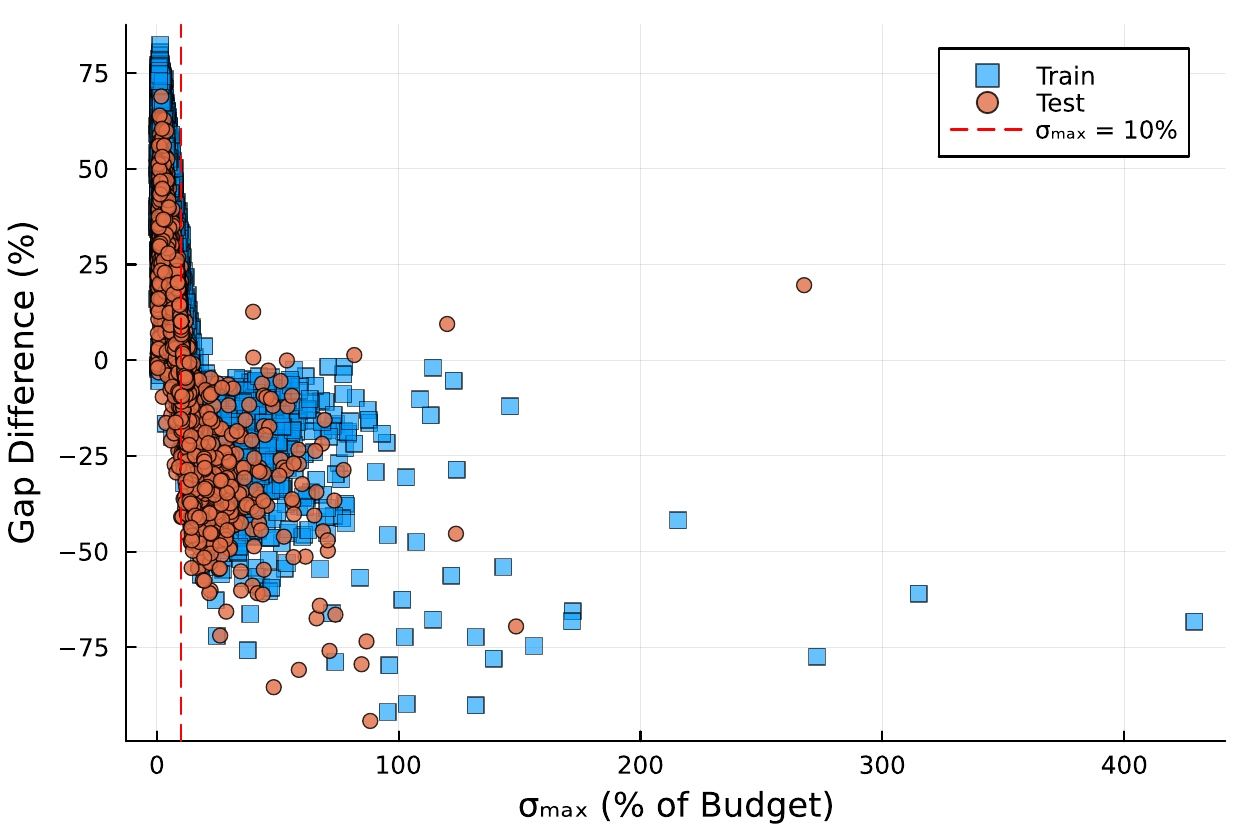}
    \captionof{figure}{Change in Relative Optimality Gap  
      ($\text{Gap}_{\text{Sobolev}}-\text{Gap}_{\text{Benchmark}}$).  
      Negative Values Indicate Improvement.}
    \label{fig:lux_gap_reduction_test}
  \end{minipage}
\end{figure}

Figure~\ref{fig:lux_gap_reduction_test} displays the gap
difference across train and test instances.  In the tight-risk regime
$\sigma_{\max}\le0.10\,\mathcal B$, where first-order information is
less informative, the Sobolev proxy under-performs the benchmark.
Outside that region, Sobolev training achieves a clear advantage.
Table~\ref{tab:gap-test-stats} provides detailed statistics; the
pronounced split suggests the benefit of a mixture-of-experts approach
that calls the benchmark in the highly constrained region and the
Sobolev proxy elsewhere.

\vspace{-0.5cm}
\section{Limitations}
\label{sec:limitations}

\vspace{-0.3cm}
\paragraph{Nested Automatic Differentiation and Jacobian Overconstraint.}
Initial experiments encountered two critical challenges: most neural–network frameworks did not efficiently support higher‐order or derivative‐based losses on GPUs, leading to training times of up to 8 hours for 6{,}000 epochs, and enforcing every entry of the solver Jacobian in each batch overconstrained the model, degrading performance relative to the MSE‐only benchmark. To address these issues, nested‐AD capabilities of \texttt{Lux.jl} were employed, dramatically shortening training times and enabling live monitoring of loss components. Additionally, random masking of 95\% of the derivative entries in each mini‐batch alleviated the overconstraint, resulting in faster convergence and improved predictive accuracy.

\vspace{-0.2cm}
\paragraph{Regularity Conditions and Sensitivity Smoothness.}
The theoretical guarantees of Sobolev Training (see Sec.~\ref{sec:background} and \ref{sec:theory}) rely on regularity conditions—such as LICQ, SOSC, and strict complementarity—to ensure a smooth, well‐defined mapping from parameters to solutions. Although many OPF instances satisfy these assumptions, occasional degeneracies (e.g., non‐unique active sets) can produce invalid or discontinuous sensitivities. Techniques such as the “fix‐and‐relax” method~\citep{pirnay2012optimal} or the corrective strategies in \citep{andersson2019casadi} can restore valid derivative information. In the conducted experiments, such irregular cases were present but did not materially affect overall proxy performance.

\vspace{-0.2cm}
\paragraph{Uninformative Sensitivities.}
Certain problem classes—especially those with stepwise or piecewise‐constant mappings from parameters to variables—naturally yield zero or uninformative sensitivities. A prime example is the Lagrangian dual variables in linear programming constraints, whose gradients vanish across broad parameter intervals. Mitigation strategies may include regularization methods or alternative loss formulations designed to handle sparse or zero‐derivative signals \citep{jungel2023learning}.

\vspace{-0.5cm}
\section{Conclusion}
\label{sec:conclusion}

This work demonstrates how Sobolev Training can serve as an effective end-to-end optimization proxy by embedding solver‐based sensitivities into the learning process. A theoretical framework under Lipschitz continuity provides performance guarantees, showing that matching both values and derivatives controls the approximation error in terms of training set density. Empirical evaluation on three large AC-OPF systems shows that Sobolev training yields lower prediction error, tighter constraint satisfaction, and \emph{far fewer extreme infeasibilities} than an MSE-only proxy; moreover, on a (semi) self-supervised mean–variance portfolio task it \emph{slashes the average optimality gap by roughly 50 \% in the medium-risk regime} while matching the baseline in tighter-risk settings.

Future research may explore the incorporation of second‐order derivative information to further improve approximation fidelity, adapt the Sobolev approach to mixed‐integer and stochastic optimization problems, and develop feasibility‐restoration or projection techniques to handle instances where solver regularity conditions are marginally violated. These extensions promise to broaden the applicability and robustness of optimization proxies in large‐scale industrial settings.

\begin{ack}
This research is partly funded by NSF award 2112533. 
The work was also funded by Los Alamos National Laboratory's Directed Research and Development project, ``Artificial Intelligence for Mission (ArtIMis)'' under U.S. DOE Contract No. DE-AC52-06NA25396.
\end{ack}

\bibliographystyle{unsrt}
\bibliography{bibliography}


\appendix


\section{AC-OPF Nomenclature}
\label{app:notation}

This appendix lists every symbol and constraint that appears in the AC optimal
power-flow formulation of Model~\ref{model:acopf}.

\subsection{Sets and Indices}
\renewcommand{\arraystretch}{1.1}
\begin{center}
\begin{tabular}{c|l}
\toprule
Symbol & Definition \\ \midrule
$\NODES$ & Set of buses (nodes) in the network, indexed by $i,j$ \\
$\EDGES$ & Set of directed branches, indexed by $(i,j)$ \\
$\mathcal{E}_i$ & Branches whose \emph{from-bus} is $i$ \\
$\mathcal{E}^R_i$ & Branches whose \emph{to-bus} is $i$ \\
\bottomrule
\end{tabular}
\end{center}

\subsection{Decision Variables}
\begin{center}
\begin{tabular}{c|l|c}
\toprule
Variable & Description & Units \\ \midrule
$\SG_i$ & Complex net power generated at bus $i$ & MVA \\
$\V_i$  & Complex voltage phasor at bus $i$      & p.u.\ or kV \\
$\SF_{ij}$ & Complex apparent power flow from $i$ to $j$ & MVA \\
\bottomrule
\end{tabular}
\end{center}

\subsection{Constants}
\begin{center}
\begin{tabular}{c|l|c}
\toprule
Constant & Description & Units \\ \midrule
$c_i(\cdot)$ & Generation cost curve at bus $i$ & \$ \\
$Y_{ij}$ & Series admittance of branch $(i,j)$ & p.u.\ \\
$Y^{c}_{ij}$ & Shunt admittance of branch $(i,j)$ & p.u.\ \\
$\underline{\VM_i},\,\overline{\VM_i}$ & Voltage magnitude limits at bus $i$ & p.u.\ or kV \\
$\underline{\SG_i},\,\overline{\SG_i}$ & Generator capability limits at bus $i$ & MVA \\
$\overline{S_{ij}}$ & Thermal limit of branch $(i,j)$ & MVA \\
\bottomrule
\end{tabular}
\end{center}

\subsection{Parameters}
\begin{center}
\begin{tabular}{c|l|c}
\toprule
Parameter & Description & Units \\ \midrule
$\SD_i$ & Forecast demand at bus $i$ & MVA \\
\bottomrule
\end{tabular}
\end{center}

\subsection{Constraint Notes for Model~\ref{model:acopf}}
\begin{enumerate}[label=C\arabic*., leftmargin=10mm]
\item \eqref{model:acopf:obj}\quad Minimise total generation cost $\sum_i c_i(\SG_i)$.
\item \eqref{model:acopf:kirchhoff}\quad Nodal power balance (Kirchhoff’s current law).
\item \eqref{model:acopf:ohm:fr}–\eqref{model:acopf:ohm:to}\quad Non-linear AC power-flow relations (Ohm’s law).
\item \eqref{model:acopf:vmbound}\quad Voltage magnitude limits.
\item \eqref{model:acopf:genbound}\quad Generator capability limits.
\item \eqref{model:acopf:thrmbound}\quad Branch thermal limits.
\end{enumerate}

\footnotetext{Both directions of a physical line are treated explicitly:
$(i,j)$ and $(j,i)$ may appear separately in $\EDGES$.}

\subsection*{Legend}
Variables:\; $\SG_i$, $\SF_{ij}$, $\V_i$ \qquad
Parameters:\; all other symbols \qquad
$|\cdot|$ = magnitude,\; ${}^{\star}$ = complex conjugate




\section{Implementation}
\label{app:imp}

All experiments are implemented in the Julia programming language \cite{julia}, which combines a high‐level, dynamic syntax with just‐in‐time compilation to deliver performance on par with statically compiled languages. Julia’s multiple‐dispatch paradigm and extensive numerical ecosystem make it particularly well‐suited for blending optimization modeling, automatic differentiation, and machine learning in a single environment.

Transmission network and generator data are ingested via \texttt{PowerModels.jl}, a domain‐specific package that parses PGLib case files and constructs corresponding \texttt{JuMP.jl} models. \texttt{JuMP.jl} serves as the algebraic modeling layer, enabling concise formulation of AC‐OPF constraints and objectives and delegating solution to state‐of‐the‐art nonlinear solvers.

Dataset generation is automated with \texttt{LearningToOptimize.jl} (L2O.jl), which samples random parameter vectors, solves each OPF instance, and records both the optimal primal variables and the associated sensitivities. Sensitivities are extracted using \texttt{DiffOpt.jl}, an extension of \texttt{JuMP.jl} that computes solver‐level derivatives via nested automatic differentiation. The resulting triples \((p_i,\,x^*(p_i),\,D x^*(p_i))\) form the training corpus for the Sobolev regression loss.

Neural‐network training was split across two frameworks. The benchmark value‐only models were implemented in \texttt{Flux.jl}, chosen for its mature support and high computational throughput. In contrast, Sobolev training was carried out in \texttt{Lux.jl}, a separate Julia deep‐learning library inspired by the Flux API (but architecturally independent) with with GPU‐compatible nested higher‐order derivatives.

\begin{table}[H]
  \centering
  \caption{Training hyper–parameters for each proxy network.}
  \label{tab:train_hparams}
  \begin{tabular}{lccccc}
    \toprule
    \textbf{System} & \textbf{Layers} & \textbf{Activation} & \textbf{Batch size} & $\bm{\Lambda}$ & \textbf{Mask sparsity (\%)}\\
    \midrule
    \pegaseS      & {[}320, 320{]}  & ReLU        & 32 & $0.14$ & 5  \\
    \ieee         & {[}320, 320{]}  & Sigmoid    & 32 & $0.30$ & 5  \\
    \rte          & {[}320, 320{]}  & ReLU        & 32 & $0.1$ & 25 \\
    Markowitz     & {[}1024, 1024{]}& LeakyReLU   & 32 & $4.35$ & 5  \\
    \bottomrule
  \end{tabular}
\end{table}

\noindent\textbf{Optimizer.} All models use \texttt{Adam} $(\eta = 0.001,\; \beta = (0.9, 0.999),\; \epsilon = 10^{-8})$.

\noindent\textbf{Resources.} Experiments are carried out on Intel(R) Xeon(R) Gold 6226 CPU @
2.70GHz machines with NVIDIA Tesla A100 GPUs on the Phoenix cluster
\cite{PACE}.

\section{Supervised Mean–Variance Portfolio Optimization}
\label{app:supervised-portfolio}

This appendix evaluates a fully supervised proxy for the mean–variance task by solving each instance offline to obtain exact portfolio weights and (where available) sensitivities.  The results—showing large mean optimality gaps and high variance even with perfect labels—underscore why the main paper relies on a self-supervised training strategy, which avoids these challenges.

\begin{figure}[H]
  \centering
  \begin{minipage}[t]{0.4\textwidth}
    \centering
    \vspace{-2.7cm}
    {\footnotesize%
    \setlength{\tabcolsep}{3.5pt}
    \begin{tabular}{lcc}
      \toprule
      GAP & Sobolev & Bench \\ \midrule
      Mean & $92.3$ & $93.41$ \\ 
      STD & $2.48$ & $2.31$ \\ 
      Maximum & $98.99$ & $99.15$ \\
      Minimum & $66.33$ & $90.14$ \\\bottomrule
    \end{tabular}}
    \vspace{0.5cm}
    \captionof{table}{Test-set gap}
    \label{tab:super-gap-test-stats}
  \end{minipage}\hfill%
  \begin{minipage}[t]{0.6\textwidth}
    \centering
    \includegraphics[width=\linewidth]{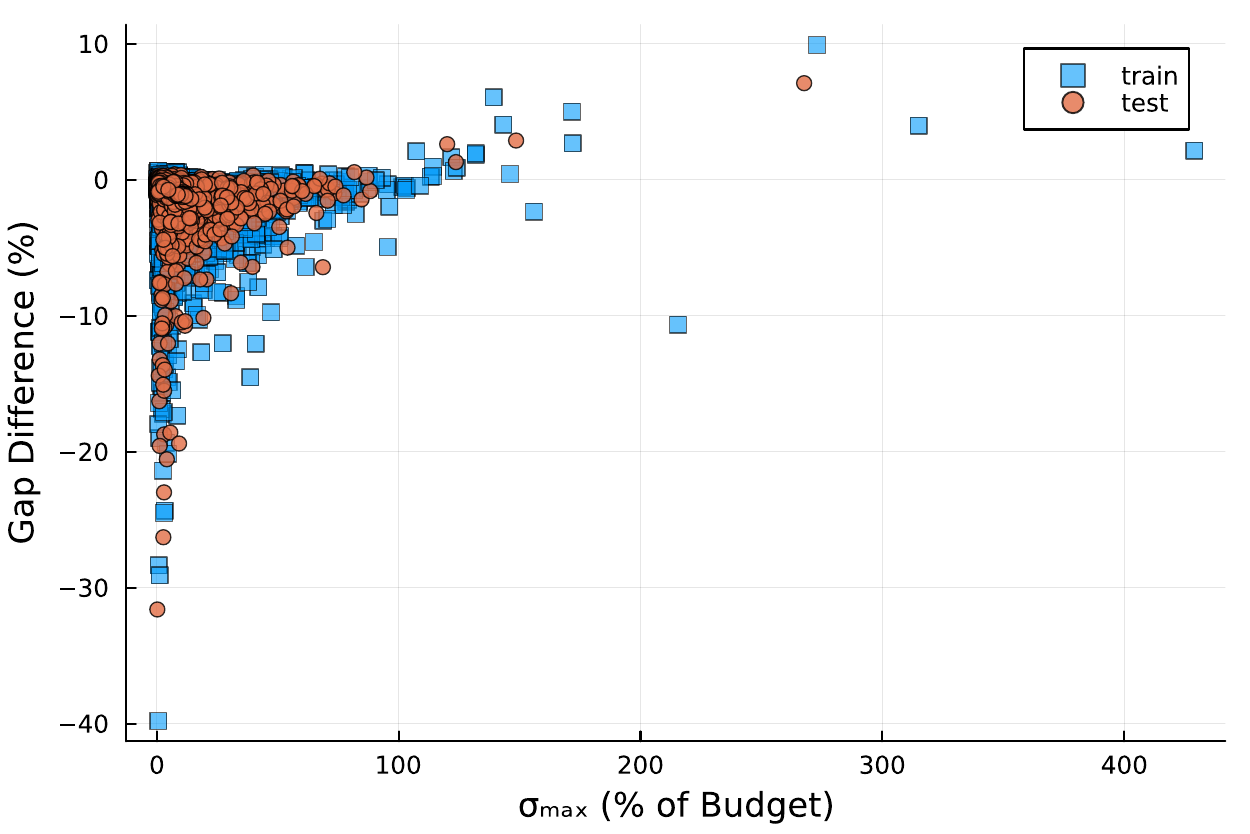}
    \captionof{figure}{Change in Relative Optimality Gap  
      ($\text{Gap}_{\text{Sobolev}}-\text{Gap}_{\text{Benchmark}}$).  
      Negative Values Indicate Improvement.}
    \label{fig:marko_gap_supervised}
  \end{minipage}
\end{figure}

\section{Mask Impact}
\label{app:mask}

As reported in Table~\ref{tab:mask_sparsity_ieee},
extreme sparsity (5–10 \%) achieves the lowest MSE.
Increasing the density beyond 25 \% degrades performance and nearly
matches the fully dense (100 \%) variant, corroborating the hypothesis
that excessive gradient information introduces contention among
constraints and hampers convergence.

\begin{table}[H]
  \centering
  \begin{tabular}{@{}cc@{}}
    \toprule
    \textbf{Mask sparsity (\%)} & \textbf{MSE} \\ \midrule
     5   & 0.0065 \\
    10   & 0.0088 \\
    25   & 0.0144 \\
   100   & 0.0148 \\ \bottomrule
  \end{tabular}
  \vspace{0.5cm}
  \caption{Impact of Jacobian‐mask sparsity on test MSE (\ieee)}
  \label{tab:mask_sparsity_ieee}
\end{table}

\end{document}